# Designing Explainable Conversational Agentic Systems for Guaraní Speakers

Samantha Adorno[1]
University of Kansas
Lawrence, Kansas, USA
samantha.adorno00@gmail.com

Akshata Kishore Moharir[2]
Independent Researcher
Sherwood, Oregon, USA
akshatankishore5@gmail.com

Ratna Kandala[1]
University of Kansas
Lawrence, Kansas, USA
ratnanirupama@gmail.com

## Abstract

Artificial intelligence systems are often presented as universal, yet their interaction paradigms remain predominantly text-first, limiting alignment with primarily oral languages and communicative practices. Using Guaraní, an official and widely spoken language of Paraguay, as a motivating case, this work examines how language support risks remaining symbolic when spoken interaction is reduced to a speech-to-text interface. We explore an oral-first, multi-agent framing in which turn-taking, repair, shared context, and governance are treated as core components of interaction rather than peripheral features. By separating language understanding from conversation state and permission mechanisms, the architecture makes conversational structure and control explicit, enabling reasoning over interaction dynamics rather than isolated commands. Framing conversational coordination as a cognitively motivated reasoning problem over shared state connects insights from human dialogue to the design of AI systems that are more interpretable and responsive in oral and low-resource settings.

## Keywords

Conversational AI, Oral-first interaction, Low-resource languages, Diglossia, Guaraní, Culturally grounded AI, Multi-Agent Systems, Indigenous data governance

## 1 Introduction

Most AI systems and everyday interaction with machines remain oriented around text input, such as keyboards, menus, and form-like interfaces. Voice features are increasingly common, but in many deployments they function primarily as a spoken front-end to a text-first pipeline (transcribe, parse, respond), rather than as conversation with turn-taking, clarification, and repair. Voice assistants such as Amazon Alexa illustrate this interaction style: wake word, short request, single response. This approach does not match the interactional structure of human dialogue, and spoken systems frequently exhibit problems with interruptions, response timing, and turn coordination [24]. When misunderstandings occur, breakdown handling often places a substantial burden on users to rephrase, repeat, or restart [1].

Conversation is a coordinated action. People manage understanding through grounding and shared context, using clarification and repair to maintain alignment [6]. Turn-taking is central to this coordination, with strong cross-cultural regularities and measurable variation in timing across languages [25]. If a system cannot manage turn-taking and repair, voice interaction tends to collapse into brittle command-and-control behavior, even when speech recognition and synthesis are available [1, 24]. These limitations disproportionately affect low-resource and endangered languages. Many such languages are primarily oral in everyday use and have limited written literature, yet language technology resources remain sparse across speech and dialogue modalities [27].

Rather than adapting oral languages to text-centric systems, oral interaction should be treated as a first-class design starting point. Orality-grounded HCI argues that design assumptions imported from literate settings can fail in oral cultures, where knowledge organization often relies on narrative structure, repetition, and socially distributed memory [23]. Recent projects demonstrate both feasibility and remaining gaps. For example, collaborations in Australia have focused on improving recognition of Aboriginal English speech varieties [12, 26]. Despite this progress, many systems continue to inherit text-first interaction assumptions, and low-resource settings often lack explicit mechanisms for dialogue state tracking, repair, and consent alongside underlying models.

We use Guaraní as a motivating case for oral-first conversational AI. Guaraní is one of Paraguay's two official languages and is widely used in daily life [19]. National statistics indicate that most speakers regularly use Guaraní, Spanish, or both [13], yet digital systems continue to privilege Spanish for interaction and disambiguation. The challenge is therefore not only recognition or translation, but supporting multi-turn interaction, shared context, and repair in the language users actually speak.

These usability failures are also explainability failures. When a system silently misinterprets an utterance, cuts off a speaker mid-turn, or recovers from a breakdown without communicating what went wrong, users cannot tell what was understood, what was inferred, or what the system intends to do next. In oral and low-resource settings, this problem is compounded because standard XAI mechanisms—written rationales, visual confidence indicators, post hoc inspection tools—are not accessible in the same way. Explainability for agentic systems must therefore be rethought as an interactional property: explanations should be spoken, timely, and embedded within dialogue rather than deferred to interfaces that assume literacy and attention outside of conversation.

To address this, we propose an oral-first assistant architecture that separates language understanding, conversation state, action execution, and governance into interacting agents. We argue that a Multi-Agent System (MAS) framework is well suited for low-resource conversational systems because it enables specialization while keeping interfaces explicit and testable. Prior work shows that decomposing conversational task-solving into specialized agents improves task success relative to monolithic models [3], and that maintaining conversational state for turn-taking and repair requires dedicated tracking distinct from generation or execution [20]. Crucially, this modular separation also supports explainability: by making interpretation, context tracking, permissions, and action decisions explicit across distinct components, the system can surface what it understood, what it plans to do, and why a given action is permitted or blocked.

## 2 Case Study: Guaraní in Paraguay

**Diglossia as a design constraint.** A central challenge for interface design in Paraguay is *diglossia*: a stable sociolinguistic arrangement in which two languages or varieties are functionally separated across domains, with a prestige hierarchy between a "High" (H) code used in formal and written contexts and a "Low" (L) code used in everyday speech [9, 10]. In Paraguay, Guaraní is widely used in oral, intimate, and community settings, while Spanish dominates literacy, bureaucracy, and public-facing written systems [15]. This division matters because most digital interaction remains mediated through written interfaces—menus, forms, error messages, and navigation flows—which implicitly assume literacy-driven reasoning and explicit symbolic input. Even when systems offer "language support," interaction often defaults to Spanish at moments requiring reading, verification, or error resolution. As a result, users may prefer Guaraní for everyday communication but rely on Spanish for institutional and digital action, reflecting a *domain mismatch* between situated reasoning and interface infrastructure [9, 15].

**Literacy and interaction burden.** Because Spanish dominates formal domains such as education and administration, written interaction frequently pulls bilingual users toward Spanish [15]. This shift is not purely linguistic: it entails a change in cognitive framing, interaction norms, and expectations of correctness and accountability. Text-based interfaces and text-first voice systems ("say it, then read or confirm it") place additional demands on explicit reasoning and self-monitoring. When breakdowns occur, users are often required to restate or correct their input in written form, reinforcing the H/L divide rather than supporting conversational repair grounded in shared context.

**Digital access vs. language choice.** Internet access in Paraguay has expanded substantially, with usage among those aged 10+ rising from 61.1% in 2017 to 81.6% in 2024 [14]. However, increased connectivity does not translate into increased Guaraní presence in digital services. Instead, mainstream tools and official channels remain optimized for Spanish literacy and explicit symbolic interaction rather than Guaraní conversational use [15]. This requires users to repeatedly shift away from everyday conversational strategies toward interfaces that privilege text-based clarification and verification. When interaction remains text-centric at critical moments such as setup, navigation, and error recovery, language support risks becoming symbolic rather than functionally integrated into users' reasoning processes.

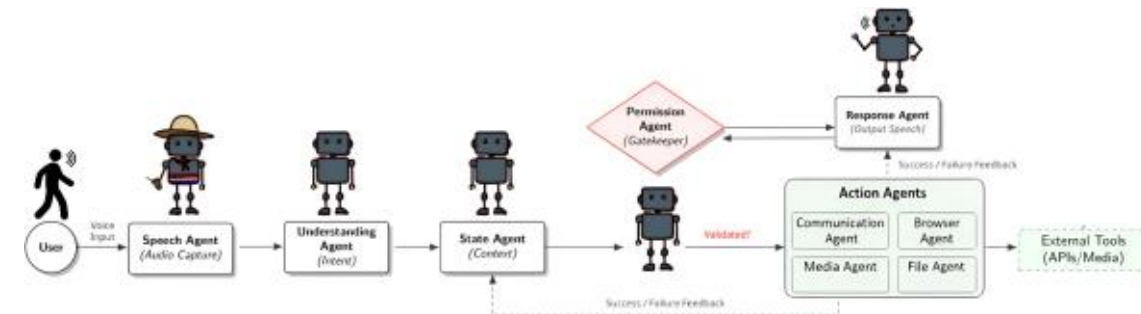

**Figure 1: Proposed Oral-First Multi-Agent Architecture. The Speech Agent (far left) is depicted in traditional Paraguayan attire to symbolize the system's grounding in local cultural identity.**

## 3 Proposed Architecture: An Oral-First Multi-Agent System

We propose an *oral-first* architecture that treats speech as the primary interaction modality, prioritizing multi-turn context, repair mechanisms, and community-led governance. Our Multi-Agent System orchestrates six specialized agents, each addressing a discrete aspect of oral language processing to transform a Guarani speaker's utterance into meaningful action:

- **Speech Interface Agent:** An agent responsible for Voice Activity Detection (VAD) and capturing audio, the first step in a conversation. In this framework, the agent is not intended to process meaning but uses pause duration and timing cues to distinguish between brief pauses and turn completion, supporting floor-holding during silence in line with spoken turn-taking behavior [24, 25]. This design choice is particularly important for Guaraní, where brief pauses or glottal stops (often referred to in pedagogical descriptions as *puso*) can occur within words and do not signal turn completion [8]. Its explainability role is to avoid silent misinterpretation of turn boundaries: when needed, it can signal to the user that the system is still listening rather than having already acted on an incomplete utterance.
- **Guaraní Understanding Agent (NLU):** With the audio captured, the Understanding agent interprets what the speaker's intentions are, mapping Guarani and Jopará speech to abstract intents like `PLAY_MUSIC`, `OPEN_TAB`. Rather than relying on synthetic translations, this agent requires training on authentic community-verified speech to recognize specific cultural references and loanwords (see section 3.2). Its explainability role is to make interpretation visible through conversational paraphrase or focused clarification rather than silently committing to an inferred intent.
- **Conversation State Agent:** Natural conversations build on shared history, and the Conversation State Agent maintains this dialogue memory. For example, if the user follows up with '*Nda che gustái*' (No, I don't like [this]), this agent resolves the implicit reference understanding that "this" refers to the current one. Its explainability role is to surface how implicit references are being resolved, allowing users to catch misunderstandings before consequences accumulate across turns.
- **Permission & Governance Agent (The "Gatekeeper"):** A sovereign agent separate from the execution layer. It checks every requested action against community-defined privacy norms (e.g., "Do not store this audio clip") and user consent settings before allowing the Action Agents to proceed. Its explainability role is central: it enables the system to tell users why an action is permitted, blocked, or subject to specific privacy constraints, so that governance decisions are never opaque system behavior.

- **Response Agent:** An agent that generates conversational responses grounded in the current dialogue state and action outcomes, including confirmations, denials, and repair prompts. All responses are checked by the Permission & Governance Agent before being delivered to the user. This agent is the primary surface through which explanations reach the user—spoken, timely, and embedded in the conversational flow rather than deferred to written interfaces.
- **Action Agents:** These are small, modular agents responsible for executing tasks in specific domains, such as a *Media Agent* (e.g., Spotify/YouTube) or a *Browser Agent*, with additional domain-specific agents introduced as needed. Their explainability role is to support concise reporting of what action was taken and, when appropriate, the reasons for failure or non-execution.

## 3.1 What Oral-First Explainability Means

Existing XAI work often assumes explanations can be read from visual displays or inspected after the fact. In oral-first settings, explanations instead need to be delivered through the flow of interaction itself. We frame oral-first explainability around four questions:

**What is being explained?** At minimum: (1) what the system understood the user to mean, (2) where uncertainty or ambiguity remains, (3) what action or tool it intends to use, (4) what permissions or governance constraints apply, and (5) what happened after execution.

**When is it explained?** Explanations may be needed *before* action (for intent confirmation or consent), *during* action (for progress or uncertainty), and *after* action (for outcome summaries, failure reasons, or repair).

**How is it explained?** Rather than relying on written rationales, the system uses conversational mechanisms: paraphrase, clarification questions, spoken previews, and repair prompts. In an agentic setting, these are not merely usability features—they are mechanisms of explainability because they make system interpretation and planned action legible to users before and after consequences occur.

**To whom is it explained?** We focus primarily on end users interacting through speech. However, the modular architecture also supports inspectability for developers and auditors, because interpretation, state tracking, permissions, and execution are separated into explicit, independently examinable components.

## 3.2 Training Data and Community Alignment

A central question for the Guaraní Understanding Agent is where training data comes from and how it can reflect authentic, community-verified speech rather than synthetic translations. Community data is essential because it captures how Guaraní speakers actually communicate, including code-switching, regional variation, and pragmatic features of oral tradition. As design reference points, we highlight two initiatives that illustrate the types of data required:

- **Mozilla Common Voice (Guaraní):** The Mozilla Common Voice Guaraní (gn) dataset is a large, crowdsourced open speech resource [2]. It illustrates how multilingual, community-contributed data can support acoustic modeling in oral-first systems.
- **The Aikuaa Initiative:** Cultural grounding can be strengthened through community-led efforts such as *Aikuaa*, a project by *El Surtidor* that organizes collaborative "mingas" to collect and validate Guaraní voice data [16]. Such initiatives capture *Jopará* usage and conversational variation that are often missing from formal or archival datasets.

| User (Guaraní) | System Action & Agent Logic |
|---|---|
| *"Che ahenduse purahei"*<br>(I want to listen to music) | 1. **Speech Agent** captures audio and detects turn completion.<br>2. **Understanding Agent** identifies intent: PLAY_MUSIC.<br>3. **Response Agent** surfaces interpretation: *"Nde rehendu purahéi resepa?"* (You want to listen to music?).<br>4. **Permission Agent** verifies "Music" is a safe category; approves.<br>5. **Media Agent** selects a popular playlist and starts playback.<br>6. **Response Agent** confirms: *"Oi˜ porã"* (Ok / It is good). |
| *"Nda che gustái"*<br>(No, I don't like [this]) | 1. **Understanding Agent** detects negative sentiment/rejection.<br>2. **Conversation State Agent** resolves the implicit object: "this" = *current song*. Updates intent to SKIP.<br>3. **Response Agent** surfaces reference resolution: *"Ndereikuaái ko purahéi?"* (You don't like this song?).<br>4. **Media Agent** executes NEXT_TRACK.<br>5. **Response Agent** confirms the new action. |

**Table 1: Multi-turn interaction demonstrating oral-first explainability through paraphrase, reference resolution, and action confirmation.**

## 3.3 Evaluation Criteria

Evaluating an oral-first architecture requires metrics beyond accuracy that capture context, sentiment, and privacy in multi-turn interaction. We consider six dimensions of conversational success:

- **Task Success Rate (TSR):** Measures the percentage of multi-turn goals completed successfully, capturing whether the Conversation State Agent maintains dialogue coherence and whether the Understanding Agent correctly interprets evolving intents across turns.
- **Repair Success Rate:** Measures conversational resilience by evaluating how often the system recovers from errors, such as misheard words or misunderstood intents, without requiring users to abandon or restart their task.

- **Explanation Comprehension:** Measures whether users correctly understood what the system interpreted, what action it planned to take, and why a given action was permitted, blocked, or revised. In oral-first settings this should be assessed through task-based probing and observational methods rather than written surveys.
- **Trust Calibration:** Evaluates whether users rely on the system appropriately rather than over-trusting fluent responses or under-trusting useful assistance. Conversational smoothness should not be taken as a proxy for correctness, and this metric tracks whether explanation mechanisms support appropriate reliance.
- **Perceived Sovereignty:** Assesses whether users trust that their voice data remains under their control. This qualitative metric evaluates confidence that audio is not stored or reused without consent and requires community-centered, ethnographic evaluation methods.
- **Latency:** Evaluates whether system response timing aligns with Guaraní conversational tempo, avoiding both premature interruptions that violate turn-taking norms and prolonged silences that disrupt conversational flow.

# 4 Discussions and Limitations

The proposed architecture highlights the potential of oral-first language technology, but several challenges extend beyond technical implementation.

## 4.1 Standardization vs. Lived Reality

A persistent challenge for Guaraní language technology is the gap between institutional standardization and everyday speech. Although Guaraní is an official language of Paraguay, supported by formal planning bodies such as the Academia de la Lengua Guaraní under the Ley de Lenguas [21, 22], daily use frequently involves *jopará* and other forms of code-switching [7, 17, 18]. As a result, the linguistically "correct" form often differs from how speakers actually communicate. Terminological resources exist, but adoption is uneven, and speakers may prefer mixed or loanword forms that preserve fluency and social meaning [21**?** ]. Oral-first systems should therefore treat variation as expected input and prioritize communicative intent over enforcing a single normative register [17, 18].

## 4.2 The Data Bottleneck is Specifically Conversational

While data scarcity is a common constraint, oral-first systems face a more specific gap: the lack of conversational audio capturing turn-taking, interruptions, repair, and shared context. Many endangered languages are primarily oral and lack extensive written resources [27]. For Guaraní, recent work continues to identify limited digital content and speech data, particularly for spontaneous interaction [5, 11]. Existing parallel corpora and scraping-based methods primarily support text-based evaluation and do not address multi-turn spoken interaction [5, 11]. Community-led data collection efforts are therefore essential, and future datasets should prioritize conversational speech over read or isolated utterances [16].

## 4.3 Governance and Perceived Control

Oral interfaces raise governance challenges because speech data is inherently identifiable and easily repurposed. Indigenous data governance frameworks emphasize community benefit, control, and accountability [4]. This motivates separating execution from a dedicated permission and privacy layer that mediates consent and data retention, including explicit "do not store audio" defaults. Such separation supports ethical commitments while increasing perceived user control and trust in multi-turn interaction [1, 4]. Critically, governance decisions must themselves be explained: a blocked action, a refusal to store audio, or a request for renewed consent should not surface as opaque system behavior. The governance layer is therefore not only a safety mechanism but also an accountability infrastructure that communicates the reasons for its constraints to users.

# 5 Conclusion

This work contributes to culturally grounded AI by treating conversation as a core computational structure rather than an interface layer. When interaction models fail to reflect how language is practiced, language support risks remaining symbolic rather than operational. Using Guaraní as a motivating case, we outline a multi-agent architecture that elevates turn-taking, repair, and shared context to first-class system components. By separating language understanding from explicit permission and governance mechanisms, the architecture makes conversational reasoning, control, and accountability inspectable and modular. This framing moves beyond universal, text-first assumptions toward interaction models that reflect human communicative coordination and sociolinguistic reality. More broadly, the work argues that equitable and aligned AI systems must reason over conversation as it is lived, particularly in oral and low-resource settings, rather than adapting those settings to inherited text-centric paradigms. A central contribution of this framing is the argument that explainability in agentic systems cannot remain tied to visual or text-heavy paradigms: in oral and diglossic settings, accountable AI must explain itself through dialogue, in context, and in the language and interactional norms users actually inhabit.

# References

[1] Essam Alghamdi, Martin Halvey, and Emma Nicol. 2024. System and User Strategies to Repair Conversational Breakdowns of Spoken Dialogue Systems: A Scoping Review. In *Proceedings of the 6th ACM Conference on Conversational User Interfaces* (Luxembourg, Luxembourg) *(CUI '24)*. Association for Computing Machinery, New York, NY, USA, Article 28, 13 pages. doi:10.1145/3640794.3665558

[2] Rosana Ardila, Megan Branson, Kelly Davis, Michael Kohler, Josh Meyer, Michael Henretty, Reuben Morais, Lindsay Saunders, Francis Tyers, and Gregor Weber. 2020. Common Voice: A Massively-Multilingual Speech Corpus. In *Proceedings of the Twelfth Language Resources and Evaluation Conference*, Nicoletta Calzolari, Frédéric Béchet, Philippe Blache, Khalid Choukri, Christopher Cieri, Thierry Declerck, Sara Goggi, Hitoshi Isahara, Bente Maegaard, Joseph Mariani, Hélène Mazo, Asuncion Moreno, Jan Odijk, and Stelios Piperidis (Eds.). European Language Resources Association, Marseille, France, 4218–4222. https://aclanthology.org/2020.lrec-1.520/

[3] Jonas Becker. 2024. Multi-Agent Large Language Models for Conversational Task-Solving. arXiv preprint arXiv:2410.22932v1. https://arxiv.org/abs/2410.22932

[4] Stephanie Russo Carroll et al. 2020. The CARE Principles for Indigenous Data Governance. *Data Science Journal* (2020).

[5] Luis Chiruzzo, Santiago Góngora, Aldo Alvarez, Gustavo Giménez-Lugo, Marvin Agüero-Torales, and Yliana Rodríguez. 2022. Jojajovai: A Parallel Guarani-Spanish Corpus for MT Benchmarking. In *Proceedings of the Thirteenth Language Resources and Evaluation Conference*, Nicoletta Calzolari, Frédéric Béchet, Philippe Blache,

Khalid Choukri, Christopher Cieri, Thierry Declerck, Sara Goggi, Hitoshi Isahara, Bente Maegaard, Joseph Mariani, Hélène Mazo, Jan Odijk, and Stelios Piperidis (Eds.). European Language Resources Association, Marseille, France, 2098–2107. https://aclanthology.org/2022.lrec-1.226/

[6] Herbert H. Clark and Susan E. Brennan. 1991. Grounding in Communication. In *Perspectives on Socially Shared Cognition*. American Psychological Association. https://www.cs.cmu.edu/~illah/CLASSDOCS/Clark91.pdf

[7] Bruno Estigarribia. 2015. Jopará and Guaraní in Paraguay (discussion of contact and mixed speech).

[8] Bruno Estigarribia. 2020. *A Grammar of Paraguayan Guaraní*. UCL Press. https://uclpress.co.uk/book/a-grammar-of-paraguayan-guarani/

[9] Charles A. Ferguson. 1959. Diglossia. *Word* 15, 2 (1959), 325–340. doi:10.1080/00437956.1959.11659702

[10] Joshua A. Fishman. 1967. Bilingualism with and without Diglossia; Diglossia with and without Bilingualism. *Journal of Social Issues* 23, 2 (1967), 29–38. doi:10.1111/j.1540-4560.1967.tb00573.x

[11] Santiago Góngora, Nicolás Giossa, and Luis Chiruzzo. 2021. Experiments on a Guarani Corpus of News and Social Media. In *Proceedings of the First Workshop on Natural Language Processing for Indigenous Languages of the Americas*, Manuel Mager, Arturo Oncevay, Annette Rios, Ivan Vladimir Meza Ruiz, Alexis Palmer, Graham Neubig, and Katharina Kann (Eds.). Association for Computational Linguistics, Online, 153–158. doi:10.18653/v1/2021.americasnlp-1.16

[12] Ben Hutchinson. 2025. A partnership with The University of Western Australia to improve speech technology for Aboriginal and Torres Strait Islander people's voices. Google Australia Blog. https://blog.google/intl/en-au/company-news/technology/a-partnership-to-improve-speech-technology-for-first-nations-voices/

[13] Instituto Nacional de Estadística (INE), Paraguay. 2024. Día Internacional de la Lengua Materna: Diversidad lingüística en Paraguay. https://www.ine.gov.py/noticias/2298/dia-internacional-de-la-lengua-materna-diversidad-linguistica-en-paraguay

[14] Instituto Nacional de Estadística (INE), Paraguay. 2025. 8 de cada 10 personas utiliza internet en Paraguay (EPH 2017–2024).

[15] H. Ito. 2012. With Spanish, Guaraní lives: a sociolinguistic analysis of bilingual education in Paraguay. *Multilingual Education* 2, 1 (2012), 6. doi:10.1186/2191-5059-2-6

[16] JournalismAI. 2025. Guarani AI: When building language tech means building community.https://www.journalismai.info/blog/5fcm6ayykhqq7564kbvt9nw92wwmy9

[17] Olga Kellert and Nemika Tyagi. 2025. Where and How Do Languages Mix? A Study of Spanish-Guaraní Code-Switching in Paraguay. In *Proceedings of the Workshop on Computational Approaches to Linguistic Code-Switching*. Association for Computational Linguistics.

[18] Katherine Mortimer. 2006. Guaraní Académico or Jopará? Educator Perspectives and Ideological Debate in Paraguayan Bilingual Education.

[19] Organization of American States (OAS). 1992. Paraguay's Constitution of 1992 with Amendments through 2011. PDF. https://www.oas.org/ext/Portals/33/Files/Member-States/Parag_intro_textfun_eng_1.pdf

[20] Sagar Sapkota et al. 2025. Multi-Party Conversational Agents: A Survey. arXiv preprint arXiv:2505.18845v1. https://arxiv.org/abs/2505.18845

[21] Secretaría de Políticas Lingüísticas. [n. d.]. Academia de la Lengua Guaraní. https://spl.gov.py/es/academia-de-la-lengua-guarani/.

[22] Secretaría de Políticas Lingüísticas (Paraguay). 2010. Ley Nº 4251/2010: Ley de Lenguas (texto bilingüe). PDF. https://spl.gov.py/files/legal/Ley%204251%20-%20bilingue.pdf

[23] Jahanzeb Sherwani, Nosheen Ali, Carolyn Penstein Rosé, and Roni Rosenfeld. 2009. Orality-Grounded HCID: Understanding the Oral User. *Information Technologies & International Development* 5, 4 (2009), 37–49. https://itidjournal.org/index.php/itid/article/download/422/422-1096-2-PB.pdf

[24] Gabriel Skantze. 2021. Turn-taking in conversational systems and human-robot interaction: A review. *Computer Speech & Language* 67 (2021), 101178. doi:10.1016/j.csl.2020.101178

[25] Tanya Stivers, N. J. Enfield, Penelope Brown, Christina Englert, Makoto Hayashi, Trine Heinemann, G. Hoymann, Federico Rossano, Jan P. de Ruiter, Kyung-Eun Yoon, and Stephen C. Levinson. 2009. Universals and cultural variation in turn-taking in conversation. *Proceedings of the National Academy of Sciences* 106, 26 (2009), 10587–10592. doi:10.1073/pnas.0903616106

[26] The University of Western Australia. 2025. First Nations people to benefit from inclusive technology partnership. UWA News. https://www.uwa.edu.au/news/article/2025/february/first-nations-people-to-benefit-from-inclusive-technology-partnership

[27] Mark Turin. 2012. Voices of vanishing worlds: Endangered languages, orality, and cognition. *Análise Social* 205, 47 (2012). https://www.researchgate.net/publication/262778986_Voices_of_vanishing_worlds_Endangered_languages_orality_and_cognition